\documentclass[11pt]{article}
\usepackage[a4paper,margin=2.5cm]{geometry}
\usepackage[T1]{fontenc}
\usepackage{cite}
\usepackage{amsmath}
\usepackage{newtxtext,newtxmath}
\usepackage{microtype}
\usepackage{graphicx}
\usepackage{textcomp}
\usepackage{xcolor}
\usepackage{booktabs}
\usepackage{float}
\usepackage[font=small,labelfont=bf]{caption}
\usepackage{authblk}
\usepackage{tikz}
\usetikzlibrary{positioning}
\usepackage[hidelinks]{hyperref}

\title{3D Oral Modelling with Improved Vertex Distribution Using Matching-Based Learning}

\author[1]{Jihun Cho}
\author[2]{Soo-Yeon Jeong}
\author[3]{Eun-Jeong Bae}
\author[1]{Sun-Young Ihm\thanks{Corresponding author.}}
\affil[1]{Dept. of Computer Engineering, Pai Chai University, Daejeon, Korea}
\affil[2]{Division of Software Engineering, Pai Chai University, Daejeon, Korea}
\affil[3]{Dept. of Dental Technology, Bucheon University, Bucheon, Korea}
\affil[ ]{\small\texttt{2161083@pcu.ac.kr}, \texttt{sy.jeong@pcu.ac.kr}, \texttt{baebae@bc.ac.kr}, \texttt{sunnyihm@pcu.ac.kr}}
\date{}

\begin{document}

\maketitle

\begin{abstract}
In our previous work, a deep learning-based framework for 3D intraoral reconstruction was proposed. The model directly predicts explicit 3D point cloud coordinates from ten fixed-angle intraoral images, employing MobileNetV2 and Multi-head Attention for multi-view feature fusion, with a combined L1 Loss and Chamfer Distance as the loss function. Although the model achieved an accuracy of 77.49\%, predicted vertices tended to concentrate in high-density regions of the ground truth, leaving other regions largely uncovered.

In this paper, an improved loss function is proposed to address this limitation. Hungarian matching with filtering and Repulsion Loss are introduced to enforce more uniform vertex distribution across the reconstructed model. The proposed model achieves an accuracy of 68.02\%, which is numerically lower than the previous model. However, the vertex clustering issue observed in the prior work is substantially alleviated, with predicted vertices distributed more evenly across the entire reconstructed surface.

\medskip
\noindent\textbf{Keywords:} 3D Reconstruction; Oral Cavity; Deep Learning; Point Cloud; Hungarian Matching; MobileNetV2
\end{abstract}

\section{Introduction}\label{sec:intro}
A prior study~\cite{cho2025korean} introduced a supervised deep learning framework that reconstructs 3D oral structures from ten fixed-angle intraoral images, reporting an accuracy of 77.49\%. Despite this result, the predicted vertices were found to cluster predominantly in high-density regions of the ground truth, leaving lower-density areas largely uncovered. This issue can be attributed to the nature of Chamfer Distance, which was used as the primary loss function and allows multiple predicted vertices to map to the same ground truth point without considering the global distribution.

To address this limitation, this paper proposes an improved loss function consisting of three components. First, Hungarian matching is applied to enforce a one-to-one correspondence between predicted and ground truth vertices, preventing duplicate mappings. Second, a filtering mechanism focuses learning on poorly matched vertex pairs, directing the model toward underrepresented regions. Third, Repulsion Loss is introduced to explicitly penalise overly close predicted vertices, encouraging a more uniform spatial distribution.

This paper is an English version of our previous work published at the Korean Multimedia Society Conference.

The remainder of this paper is organised as follows. Section~\ref{sec:related} reviews related work. Section~\ref{sec:method} describes the proposed method. Section~\ref{sec:experiments} presents experimental results and analysis. Section~\ref{sec:conclusion} concludes the paper.

\section{Related Work}\label{sec:related}
For a comprehensive review of related work on multi-view 3D reconstruction, dental 3D reconstruction, and multi-view feature fusion, we refer the reader to our previous work~\cite{cho2025korean}.

\subsection{Point Cloud Learning}
PointNet~\cite{qi2017pointnet} pioneered deep learning directly on point sets, demonstrating effective 3D shape understanding without voxelisation. Its successor, PointNet++~\cite{qi2017pointnet++}, introduced hierarchical feature learning using Farthest Point Sampling, which is adopted in this paper for vertex reduction during preprocessing.

\subsection{Hungarian Matching in Deep Learning}
Hungarian matching has been widely adopted in deep learning for set prediction tasks requiring one-to-one correspondence. DETR~\cite{carion2020detr} introduced bipartite matching via the Hungarian algorithm to assign predicted objects to ground truth targets without duplicate assignments, enabling end-to-end object detection. Inspired by this, the proposed method applies Hungarian matching to enforce one-to-one correspondence between predicted and ground truth vertices, preventing multiple predictions from clustering around the same target point.

\section{Method}\label{sec:method}

\subsection{Dataset}
\begin{figure}[htbp]
\centering
\includegraphics[width=0.9\linewidth]{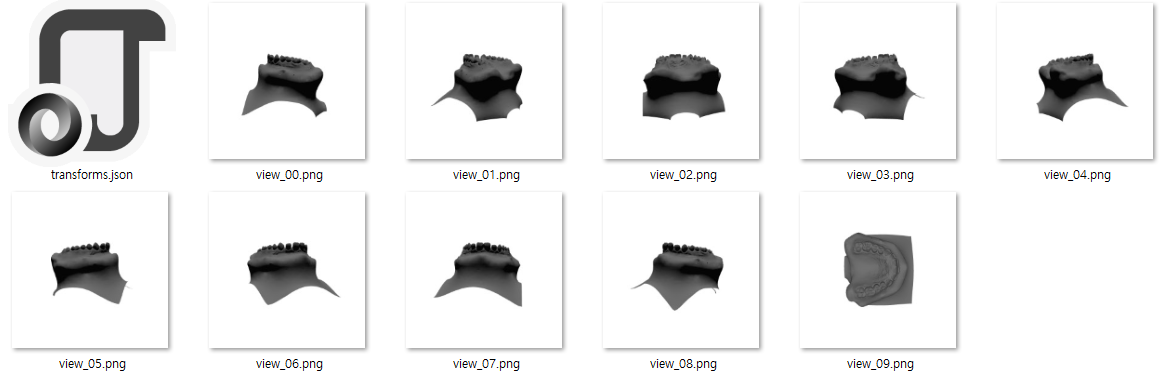}
\caption{Ten fixed viewpoints rendered from a preprocessed upper jaw mesh.}
\label{fig:views}
\end{figure}

The dataset used in this paper is the publicly available Teeth3DS dataset~\cite{teeth3ds}, consisting of 950 upper jaw samples. The preprocessing pipeline follows that of our previous work~\cite{cho2025korean}, applying centre alignment, scale normalisation, Farthest Point Sampling (FPS)~\cite{qi2017pointnet++}, Poisson surface reconstruction~\cite{kazhdan2006poisson}, and rendering from ten fixed viewpoints. Example rendered views are shown in Fig.~\ref{fig:views}.

In this paper, the number of vertices is reduced from 50,000 to 25,000. This reduction is motivated by the computational cost of Hungarian matching, which scales cubically with the number of matched points, making full 50,000-vertex matching infeasible within available GPU memory constraints.

\subsection{Model Architecture}
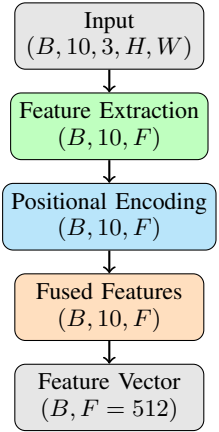
\begin{figure}[htbp]
\centering
\begin{tikzpicture}[
    node distance=0.4cm,
    box/.style={rectangle, rounded corners, draw, minimum width=2.3cm, minimum height=1.0cm, align=center, font=\small},
]
\node[box, fill=gray!20] (input) {Input\\$(B, 10, 3, H, W)$};
\node[box, fill=green!25, right=of input] (feat) {Feature\\Extraction\\$(B, 10, F)$};
\node[box, fill=cyan!25, right=of feat] (pos) {Positional\\Encoding\\$(B, 10, F)$};
\node[box, fill=orange!25, right=of pos] (fused) {Fused Features\\$(B, 10, F)$};
\node[box, fill=gray!20, right=of fused] (out) {Feature Vector\\$(B, F=512)$};
\draw[->, thick] (input) -- (feat);
\draw[->, thick] (feat) -- (pos);
\draw[->, thick] (pos) -- (fused);
\draw[->, thick] (fused) -- (out);
\end{tikzpicture}
\caption{Encoder architecture.}
\label{fig:encoder}
\end{figure}

\begin{figure}[htbp]
\centering
\begin{tikzpicture}[
    node distance=0.4cm,
    box/.style={rectangle, rounded corners, draw, minimum width=2.3cm, minimum height=1.0cm, align=center, font=\small},
]
\node[box, fill=gray!20] (input) {Input\\$(B, 512)$};
\node[box, fill=green!25, right=of input] (mod) {Feature\\Modulator};
\node[box, fill=cyan!25, right=of mod] (mlp) {MLP\\$512{\to}1024{\to}1024{\to}75{,}000$};
\node[box, fill=orange!25, right=of mlp] (reshape) {Reshape\\$(B, 25000, 3)$};
\draw[->, thick] (input) -- (mod);
\draw[->, thick] (mod) -- (mlp);
\draw[->, thick] (mlp) -- (reshape);
\end{tikzpicture}
\caption{Decoder architecture.}
\label{fig:decoder}
\end{figure}
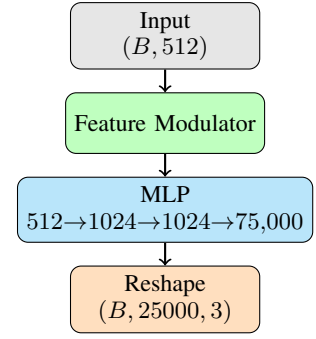

The proposed model follows the same encoder-decoder architecture as our previous work~\cite{cho2025korean}. The encoder takes ten fixed-angle intraoral images as input and extracts multi-view features using MobileNetV2~\cite{sandler2018mobilenetv2} pretrained on ImageNet~\cite{deng2009imagenet}, which are fused via positional embeddings and multi-head attention~\cite{vaswani2017attention} to produce a single feature vector of shape $(B, 512)$. The decoder then expands this feature vector through an MLP with Dropout~\cite{srivastava2014dropout} for regularisation to directly predict 25,000 3D vertex coordinates, outputting a tensor of shape $(B, 25000, 3)$. The encoder and decoder architectures are illustrated in Fig.~\ref{fig:encoder} and Fig.~\ref{fig:decoder} respectively.

\subsection{Loss Function}
The loss function proposed in this paper consists of three components: Hungarian matching loss with filtering, L1 Loss, and Repulsion Loss. The total loss is defined as:

\begin{equation}
\mathcal{L} = \alpha \cdot \mathcal{L}_{matched} + \beta \cdot \mathcal{L}_{L1} + \gamma \cdot \mathcal{L}_{repulsion}
\label{eq:loss}
\end{equation}

where $\alpha$, $\beta$, and $\gamma$ are epoch-wise weights that are scheduled throughout training.

\subsubsection{Motivation}
In our previous work~\cite{cho2025korean}, Chamfer Distance was used as the primary loss function. Chamfer Distance computes the nearest-neighbour distance from each predicted vertex to the ground truth, and vice versa. While effective at minimising overall positional error, this approach allows multiple predicted vertices to map to the same ground truth vertex, as each prediction is matched independently without considering the global distribution. As a result, predicted vertices tend to cluster in high-density regions of the ground truth, where minimising the nearest-neighbour distance is easiest, leaving other regions largely uncovered.

To address this, the proposed loss function introduces three components designed to enforce a more uniform vertex distribution: Hungarian matching with filtering, and Repulsion Loss.

\subsubsection{Hungarian Matching}
To prevent multiple predicted vertices from mapping to the same ground truth vertex, Hungarian matching~\cite{carion2020detr} is applied to enforce a strict one-to-one correspondence. Given a sampled set of 5,000 predicted vertices and 5,000 ground truth vertices, a pairwise distance matrix is computed, and the Hungarian algorithm finds the optimal assignment that minimises the total matching cost:

\begin{equation}
\sigma^* = \arg\min_{\sigma} \sum_{i=1}^{N} \| p_i - g_{\sigma(i)} \|_2
\label{eq:hungarian}
\end{equation}

where $p_i$ denotes the $i$-th predicted vertex, $g_{\sigma(i)}$ denotes the matched ground truth vertex under permutation $\sigma$, and $N=5{,}000$ is the number of sampled vertices. This ensures that each predicted vertex is uniquely assigned to a distinct ground truth vertex, eliminating the duplicate mappings that caused clustering in the previous approach.

\subsubsection{Filtering}
Following Hungarian matching, a filtering mechanism is applied to focus learning on poorly matched vertex pairs. For each matched pair $(p_i, g_{\sigma^*(i)})$, the distance is compared against a dynamic threshold $\tau$:

\begin{equation}
\tau = 0.035 \times \left(1 - 0.7 \times \frac{t}{T}\right)
\label{eq:threshold}
\end{equation}

where $t$ is the current epoch and $T$ is the total number of epochs. Only pairs with distance exceeding $\tau$ contribute to the loss:

\begin{equation}
\mathcal{L}_{matched} = \frac{1}{|\mathcal{B}|} \sum_{i \in \mathcal{B}} \| p_i - g_{\sigma^*(i)} \|_2
\label{eq:matched_loss}
\end{equation}

where $\mathcal{B} = \{i : \| p_i - g_{\sigma^*(i)} \|_2 > \tau \}$ is the set of poorly matched pairs. The threshold $\tau$ decreases as training progresses, becoming more stringent over time. In early epochs, a higher threshold allows more vertex pairs to contribute to the loss, enabling broad coverage of the reconstructed surface. As training proceeds, the threshold is gradually reduced, focusing the loss increasingly on the remaining poorly matched regions.

\subsubsection{Repulsion Loss}
While Hungarian matching and filtering improve the global distribution of predicted vertices, they do not explicitly constrain the distances between predicted vertices themselves. To address this, Repulsion Loss is introduced to penalise predicted vertices that are too close to one another. For each predicted vertex, the distances to its $k=5$ nearest neighbours among the predicted vertices are computed, and a penalty is applied whenever the distance falls below a minimum threshold of $d_{min}=0.015$. This encourages predicted vertices to maintain a minimum spatial separation, promoting a more uniform distribution across the reconstructed surface.

\subsubsection{Total Loss}
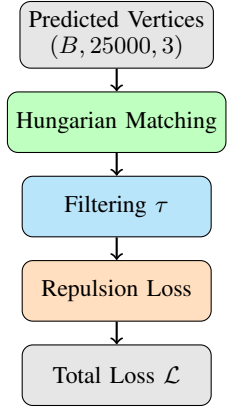
\begin{figure}[htbp]
\centering
\begin{tikzpicture}[
    node distance=0.4cm,
    box/.style={rectangle, rounded corners, draw, minimum width=2.3cm, minimum height=1.0cm, align=center, font=\small},
]
\node[box, fill=gray!20] (input) {Predicted Vertices\\$(B, 25000, 3)$};
\node[box, fill=green!25, right=of input] (hungarian) {Hungarian\\Matching};
\node[box, fill=cyan!25, right=of hungarian] (filtering) {Filtering $\tau$};
\node[box, fill=orange!25, right=of filtering] (repulsion) {Repulsion\\Loss};
\node[box, fill=gray!20, right=of repulsion] (loss) {Total Loss $\mathcal{L}$};
\draw[->, thick] (input) -- (hungarian);
\draw[->, thick] (hungarian) -- (filtering);
\draw[->, thick] (filtering) -- (repulsion);
\draw[->, thick] (repulsion) -- (loss);
\end{tikzpicture}
\caption{Loss function pipeline.}
\label{fig:loss_pipeline}
\end{figure}

The total loss is defined in Equation~\ref{eq:loss}, combining the three components described above. The epoch-wise weight scheduling is summarised in Table~\ref{tab:weight_schedule}, and the loss function pipeline is illustrated in Fig.~\ref{fig:loss_pipeline}. In the early stages of training, higher weight is assigned to L1 Loss to establish coarse positional alignment. As training progresses, the weight of the matched filtered loss is increased to refine vertex distribution, while Repulsion Loss maintains a consistent weight throughout to continuously enforce spatial separation between predicted vertices.

\begin{table}[htbp]
\centering
\caption{Loss Weight Scheduling by Epoch}
\label{tab:weight_schedule}
\begin{tabular}{cccc}
\toprule
\textbf{Epoch} & \textbf{$\alpha$ (Matched)} & \textbf{$\beta$ (L1)} & \textbf{$\gamma$ (Repulsion)} \\
\midrule
0 -- 9   & 0.5 & 0.4 & 0.1 \\
10 -- 24 & 0.7 & 0.2 & 0.1 \\
25+      & 0.8 & 0.1 & 0.1 \\
\bottomrule
\end{tabular}
\end{table}

\section{Experiments}\label{sec:experiments}

\subsection{Experimental Setup}
All experiments were conducted on a single NVIDIA RTX 5070 GPU. The detailed training configuration is summarised in Table~\ref{tab:setup}.

\begin{table}[htbp]
\centering
\caption{Experimental Setup}
\label{tab:setup}
\begin{tabular}{ll}
\toprule
\textbf{Configuration} & \textbf{Value} \\
\midrule
GPU & NVIDIA RTX 5070 (12GB VRAM) \\
Python & 3.12.9 \\
PyTorch & 2.8.0 \\
CUDA & 12.8 \\
Optimizer & AdamW~\cite{loshchilov2019adamw} \\
Initial learning rate & 0.001 \\
Weight decay & $1 \times 10^{-4}$ \\
LR scheduler & CosineAnnealingLR ($T_{max}=30$) \\
Batch size & 2 \\
Epochs & 50 \\
Early stopping patience & 15 \\
Vertex count & 25,000 \\
Hungarian sample size & 5,000 \\
\bottomrule
\end{tabular}
\end{table}

\subsection{Results}
\begin{figure}[htbp]
\centering
\includegraphics[width=0.8\linewidth]{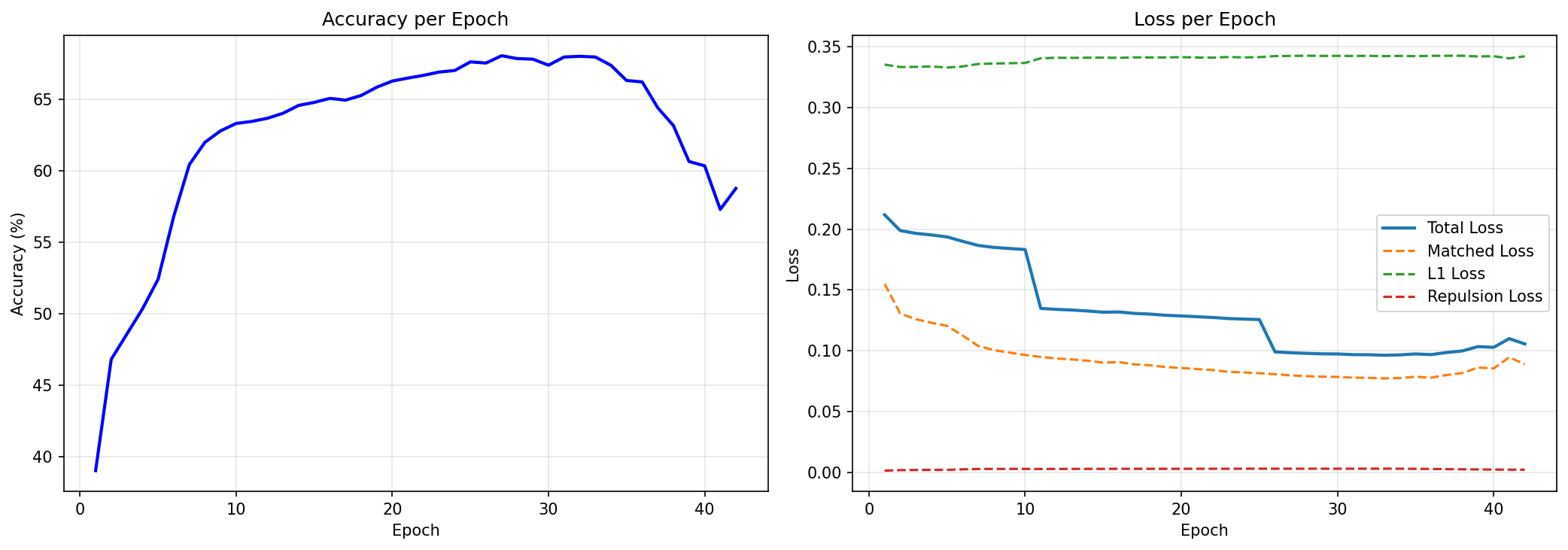}
\caption{Training curves showing accuracy and loss per epoch.}
\label{fig:training_curve}
\end{figure}

\begin{figure}[htbp]
\centering
\includegraphics[width=0.8\linewidth]{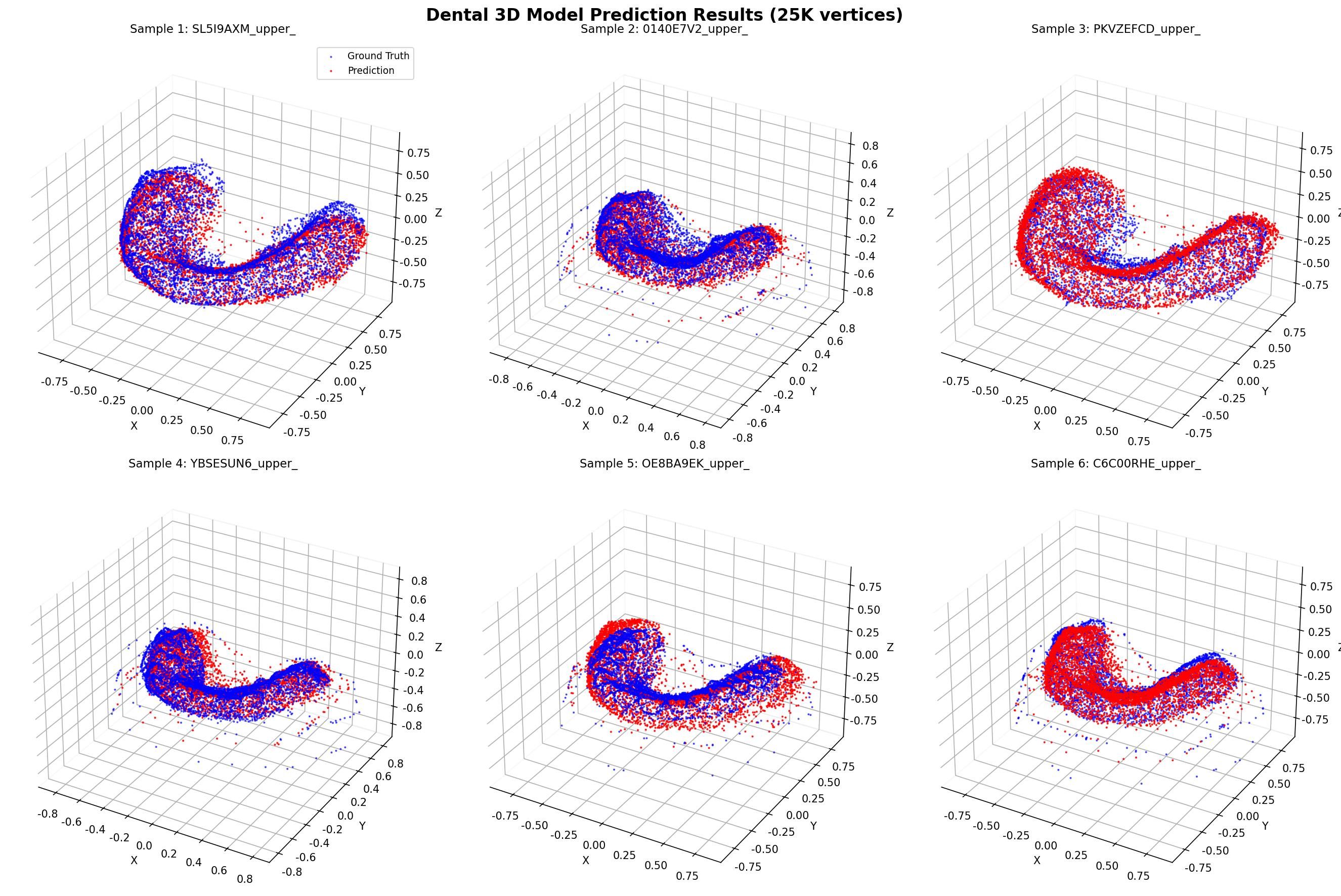}
\caption{Prediction results on six upper jaw samples. Blue: Ground Truth, Red: Prediction.}
\label{fig:results}
\end{figure}

The proposed model achieves an accuracy of 68.02\%, converging at epoch 27 of 50. The final Total Loss was 0.098291, with a Matched Filtered Loss of 0.079689, an L1 Loss of 0.342562, and a Repulsion Loss of 0.002831. While the accuracy is numerically lower than the 77.49\% reported in our previous work~\cite{cho2025korean}, the vertex distribution across the reconstructed surface is substantially improved. The training curves are shown in Fig.~\ref{fig:training_curve}, and the prediction results are visualised in Fig.~\ref{fig:results}.

\subsection{Analysis}
The proposed method achieves a lower accuracy of 68.02\% compared to 77.49\% reported in the previous work~\cite{cho2025korean}. However, this numerical decrease does not indicate a degradation in reconstruction quality. In the previous work, predicted vertices concentrated in high-density regions of the ground truth, artificially inflating the accuracy metric. Since the accuracy is measured by the proportion of predicted vertices within a fixed distance threshold of the ground truth, clustering in dense regions naturally yields higher scores. In contrast, the proposed method distributes vertices more uniformly across the reconstructed surface, which inevitably results in a lower but more meaningful accuracy.

To further validate this, a one-to-one Hungarian matching accuracy is computed, which enforces a strict bijective correspondence between predicted and ground truth vertices, preventing any single ground truth vertex from being matched more than once. Hungarian matching accuracy is evaluated by randomly sampling 5,000 vertices from both predicted and ground truth point sets, computing the optimal one-to-one assignment, and measuring the proportion of matched pairs within a distance threshold of 0.035. This procedure is repeated five times per sample and averaged to reduce variance. As Hungarian matching scales cubically with the number of points, evaluating the full 25,000 vertices at once would require constructing a $25{,}000 \times 25{,}000$ distance matrix, which exceeds available GPU memory constraints. This sampling approach therefore underestimates the true accuracy. Despite this constraint, the proposed method achieves a Hungarian accuracy of 35.71\%, compared to 8.61\% for the previous work, as summarised in Table~\ref{tab:hungarian}.

\begin{table}[htbp]
\centering
\caption{Per-sample Hungarian Matching Accuracy Comparison}
\label{tab:hungarian}
\begin{tabular}{lcc}
\toprule
\textbf{Sample} & \textbf{Previous Work} & \textbf{Proposed Method} \\
\midrule
Sample 1  & 9.64\%  & 44.92\% \\
Sample 2  & 6.27\%  & 34.84\% \\
Sample 3  & 9.96\%  & 22.04\% \\
Sample 4  & 11.52\% & 50.48\% \\
Sample 5  & 9.50\%  & 44.98\% \\
Sample 6  & 8.47\%  & 45.84\% \\
Sample 7  & 11.63\% & 31.22\% \\
Sample 8  & 3.32\%  & 15.66\% \\
Sample 9  & 9.68\%  & 29.36\% \\
Sample 10 & 6.11\%  & 37.74\% \\
\midrule
\textbf{Average} & \textbf{8.61\%} & \textbf{35.71\%} \\
\bottomrule
\end{tabular}
\end{table}

\begin{figure}[htbp]
\centering
\includegraphics[width=0.8\linewidth]{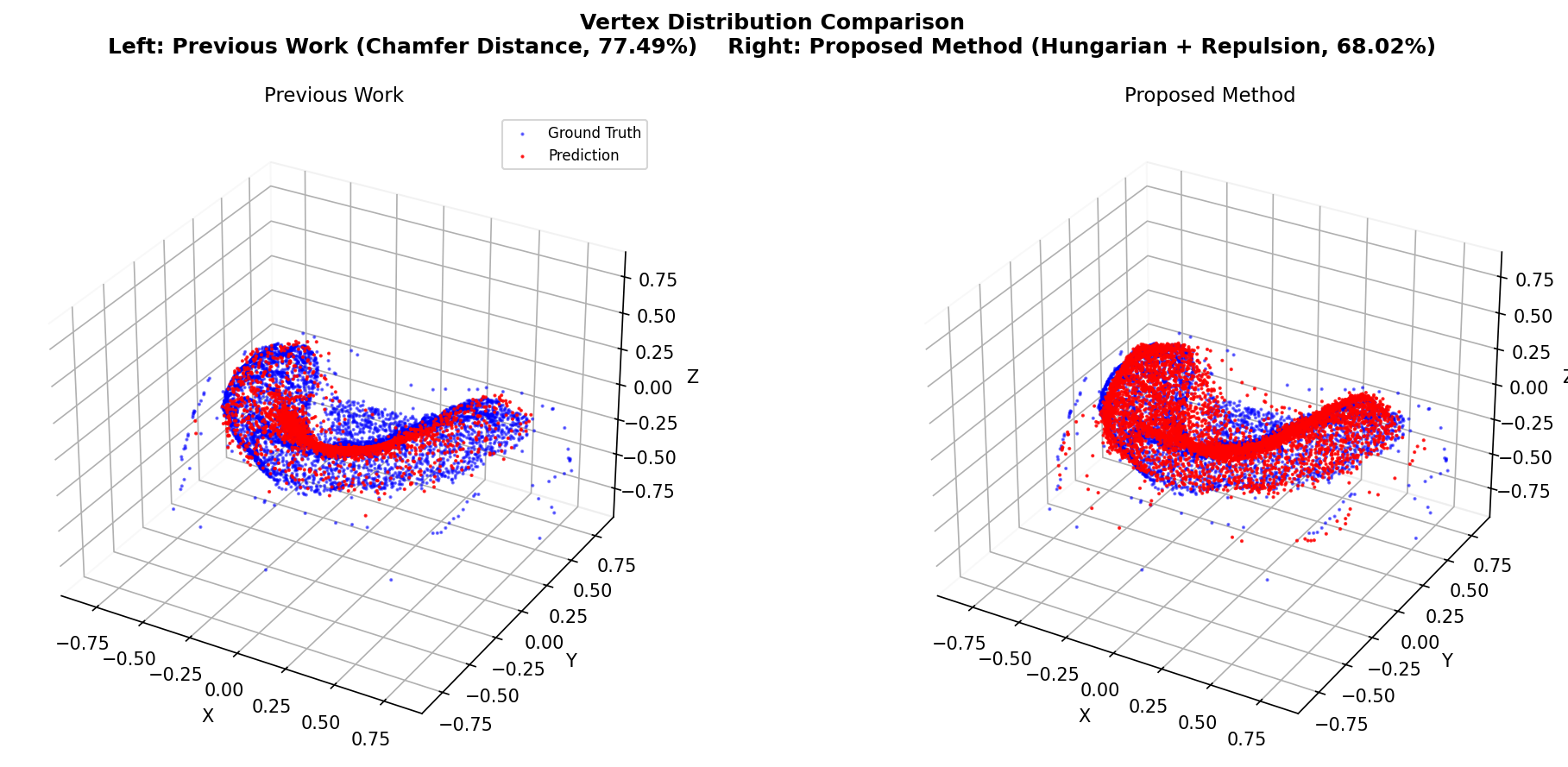}
\caption{Vertex distribution comparison between previous work (left) and proposed method (right). Blue: Ground Truth, Red: Prediction.}
\label{fig:comparison}
\end{figure}

As shown in Fig.~\ref{fig:comparison}, the previous work produces predictions that cluster predominantly in certain regions, while the proposed method achieves a more even distribution across the entire upper jaw structure.

Nevertheless, the proposed method has a notable limitation. Due to the cubic computational complexity of Hungarian matching, only 5,000 vertices are sampled per training step rather than the full 25,000. This partial matching reduces the effectiveness of the one-to-one correspondence constraint. Addressing this limitation would require either more computationally efficient matching algorithms or higher-capacity hardware capable of processing the full vertex set.

\section{Conclusion}\label{sec:conclusion}
This paper proposes an improved loss function for 3D oral cavity reconstruction, addressing the vertex clustering limitation identified in our previous work~\cite{cho2025korean}. The proposed loss function combines Hungarian matching with filtering and Repulsion Loss to enforce a more uniform vertex distribution across the reconstructed upper jaw surface.

The proposed model achieves an accuracy of 68.02\%, which is numerically lower than the 77.49\% reported in the previous work. However, this decrease is attributed to the nature of the conventional accuracy metric, which rewards vertex clustering in high-density regions. Under the one-to-one Hungarian matching accuracy metric, the proposed method achieves 35.71\% compared to 8.61\% for the previous work, demonstrating substantially more uniform vertex distribution.

A key limitation of the proposed method is that Hungarian matching is performed on 5,000 sampled vertices per training step due to GPU memory constraints, rather than the full 25,000. This reduces the effectiveness of the one-to-one correspondence constraint and likely limits the achievable accuracy. Future work will explore more computationally efficient matching algorithms or higher-capacity hardware to enable full-vertex matching, which is expected to further improve reconstruction quality and vertex distribution uniformity.

\section*{Acknowledgment}
This work was supported by the Institute of Information \& Communications Technology Planning \& Evaluation (IITP) grant funded by the Korea government (MSIT) (IITP-2026-RS-2022-00156334).

\end{document}